\documentclass{article}

\usepackage[utf8]{inputenc} 
\usepackage[T1]{fontenc}    
\usepackage{hyperref}       
\usepackage{url}            
\usepackage{booktabs}       
\usepackage{amsfonts}       
\usepackage{nicefrac}       
\usepackage{microtype}      
\usepackage{graphicx}
\usepackage{geometry}
\usepackage{upgreek}
\usepackage{multibib}

\title{Precise neural network computation \\ with imprecise analog devices}

\author{%
  Jonathan Binas, Danny Neil, \\
  Giacomo Indiveri, Shih-Chii Liu, Michael Pfeiffer \\
  \texttt{jbinas@gmail.com}
}

\begin{document}

\maketitle

\begin{abstract}
The operations used for neural network computation map favorably onto simple analog circuits, which outshine their digital counterparts in terms of compactness and efficiency.
Nevertheless, such implementations have been largely supplanted by digital designs, partly because of device mismatch effects due to material and fabrication imperfections.
We propose a framework that exploits the power of deep learning to compensate for this mismatch by incorporating the measured device variations as constraints in the neural network training process.
This eliminates the need for mismatch minimization strategies and allows circuit complexity and power-consumption to be reduced to a minimum.
Our results, based on large-scale simulations as well as a prototype VLSI chip implementation indicate a processing efficiency comparable to current state-of-art digital implementations.
This method is suitable for future technology based on nanodevices with large variability, such as memristive arrays.
\end{abstract}

\vskip 1em

 The growing need for computing power has led to the exploration of computing technologies beyond the predominant von Neumann architecture.
 In particular, due to the separation of memory and processing elements, traditional computing systems experience a bottleneck when dealing with problems involving great amounts of high-dimensional data \cite{backus1978,indiveri2015memory}, such as image processing, probabilistic inference, or speech recognition.
 These problems are often best tackled by conceptually simple but powerful and highly parallel models, such as deep neural networks (DNNs), which have delivered state-of-the-art performance on exactly those applications \cite{lecun2015deep}.
 The fact that DNNs are characterized by stereotypical and simple operations at each unit, which often can be performed in parallel, makes them compatible with the processing style of graphics processing units (GPUs) \cite{scherer2010accelerating}.
 The large computational demands of DNNs have simultaneously sparked interest in methods that make neural network inference faster and more power efficient, whether through new algorithmic inventions \cite{han2015deep, hinton2015distilling, courbariaux2015binaryconnect}, dedicated digital hardware implementations \cite{cavigelli2015origami,gokhale2014240,chen2014dadiannao,moons2018binareye}, or by taking inspiration from real nervous systems \cite{farabet2012comparison,oconnor2013real,merolla668,indiveri2015neuromorphic,pei2019towards}.

 With synchronous digital logic being the established standard of the electronics industry, many attempts towards hardware deep network accelerators have focused on this approach \cite{cavigelli2015origami,griffin199111,chen201614,park2016energy}.
 However, the massively parallel style of computation in neural networks is not reflected in the mostly serial and time-multiplexed nature of digital systems.
 An arguably more natural way of building a hardware neural network emulator is to implement its computational primitives as massively parallel physical instances of analog computing nodes, where memory and processing elements are co-localized, and state variables are directly represented by analog currents or voltages, rather than being encoded digitally \cite{rosenblatt1958,Alspector_Allen87,Vittoz90,Borgstrom_etal90,andreou1991current,Satyanarayana_etal92}.
 By directly representing neural network operations in the physical properties of silicon transistors, such analog implementations can outshine their digital counterparts in terms of simplicity, allowing for significant advances in speed, size, and power consumption \cite{hasler2013finding,masa1994high}.
 For instance, when representing quantities as currents, addition is implemented by simply connecting together wires; multiplication by a constant can be implemented with as few as two transistors. This is in stark contrast to digital implementations, where hundreds or thousands of transistors are required for a single multiplier circuit.
 The main reason why engineers have been discouraged from following this approach is that the properties of analog circuits are affected by the physical imperfections inherent to any chip fabrication process, which can lead to significant functional differences between individual devices \cite{Pelgrom_1989}.
 
Our work proposes a new approach, whereby rather than brute-force engineering more homogeneous circuits (e.g. by increasing transistor sizes and burning more power), we employ neural network training methods as an effective optimization framework to automatically compensate for the  device mismatch effects  of analog VLSI circuits.
We use measured response characteristics of individual VLSI devices as constraints in an off-line training process, to yield network configurations that are tailored to the particular analog device used, thereby compensating the inherent variability of chip fabrication.

In addition to introducing a joint training method for both device and network, we also propose compact and low-power candidate VLSI circuits.
A closed-loop demonstration of the framework is shown, based on a fabricated prototype chip, as well as detailed, large-scale simulations.
The resulting analog electronic neural network performs as well as an ideal network, while offering lower power consumption over its digital counterpart.

\begin{figure}[tp]
	\centering
	\fbox{\includegraphics[width=0.63\textwidth]{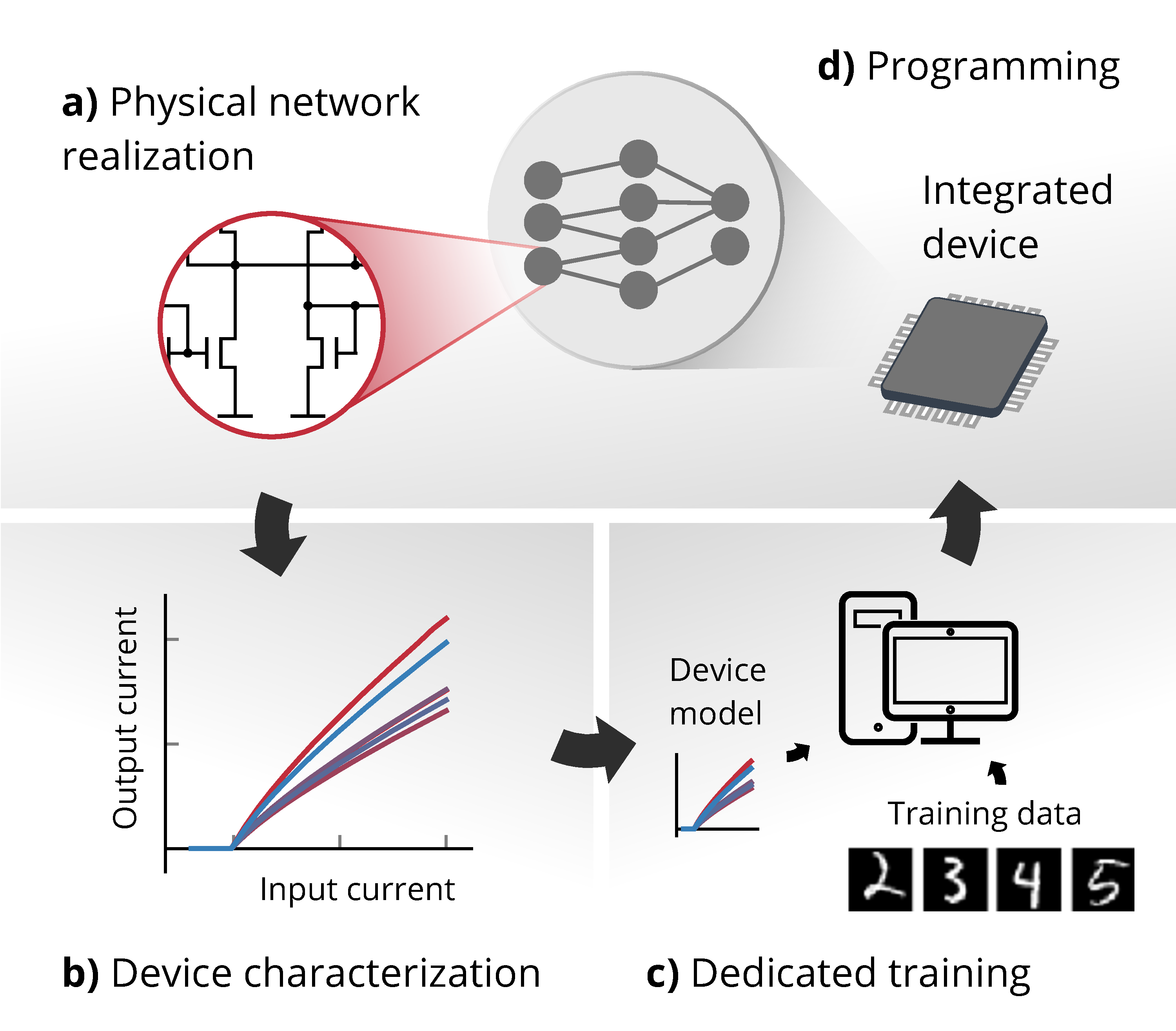}}
	\caption{Implementing and training analog electronic neural networks.
 a) The configurable network is realized on a physical substrate by means of analog circuits, together with local memory elements that store the weight configuration.
 b) The transfer characteristics of individual neurons are measured by applying specific stimuli to the input layer and simultaneously recording the output of the network. Repeating these measurements for different weight configurations and input patterns allows to reconstruct the individual transfer curves and fit them by a model to be used for training.
 c) Including the measured transfer characteristics in the training process allows optimization of the network for the particular device that has been measured.
 d) Mapping the parameters found by the training algorithm back to the device implements a neural network, whose overall computation is comparable to the theoretically ideal network.
 Arrows indicate the sequence of steps taken as well as the flow of measurement/programming data.
	}
	\label{fig:cartoon_method}
\end{figure}

\section{Related work}
 
Analog electronic implementations are almost as old as artificial neural networks themselves, with Rosenblatt's perceptron first realized as a manually configured analog circuit \cite{rosenblatt1958,hay1960mark}.
Numerous approaches have been proposed ever since, including circuits with built-in learning capabilities \cite{Alspector_Allen87}, artificially slowed-down computation \cite{Vittoz90}, bidirectional  \cite{Borgstrom_etal90,andreou1991current}, or reconfigurable connectivity \cite{Satyanarayana_etal92}.
These implementations normally do not embrace the mismatch, but rather try to compensate for it with stabilization circuits or more robust (but less efficient) circuit elements.
Moreover, these implementations typically make use of explicit multiplier circuits and require analog signals encoding the weight value.
We propose circuits that are both simpler than previous ones and at the same time can be modeled more accurately.
By not requiring different instances of a circuit to behave exactly the same, we are able to reduce them to a few essential components, and thereby achieve more compact, and arguably more elegant realizations than previous approaches.
By using digital memory elements for the network parameters and co-locating them with analog processing elements we combine the best of both worlds: we get rid of the memory access overhead of fully digital systems, which due to more complex circuits need to multiplex processing elements, but we retain reliable storage.

A number of recent approaches capitalize on the presence of mismatch in transistor circuits.
The Neural Engineering Framework (NEF) \cite{eliasmith2004neural}, for instance, requires neurons with a wide range of transfer functions of different slopes and intercepts to encode signals in the dynamics of populations of neurons. The framework has been used in various studies to implement dynamical control systems based on spiking neuron dynamics \cite{choudhary2012silicon,mundy2015efficient,voelker2017extending}.
The use of mismatch has also been key to a class of models called Extreme Learning Machines (ELMs), where the variability in neuron responses is used to construct a family of nonlinear functions. These approaches have been explored with integrated circuit implementations~\cite{chen2016elm,thakur2016elm}. 
Other than ours, these approaches rely on a certain mismatch distribution, and standard machine learning models cannot directly be mapped onto these systems.


\section{Analog electronic implementation}
\label{sect:analog}

 To illustrate our approach, we consider a multilayer perceptron, where the output of a neuron $i$ in layer $l$ is given by $y_i^l = f\smash{(\sum_j w_{ij} y_j^{l-1})}$, where $f$ is a nonlinearity, and $y^0 \equiv x$ is the input signal.
 The basic operations comprising such a system -- summation, multiplication by scalars, and simple nonlinear transformations -- can be implemented in analog electronic circuitry very efficiently, that is with very few transistors, whereby numeric values are represented by actual voltage or current values, rather than a digital code.
 Analog circuits are affected by fabrication mismatch, i.e. small fluctuations in the fabrication process that lead to fixed distortions of functional properties of elements on the same device, as well as multiple sources of noise.
 As a consequence, the response of an analog hardware neuron is slightly different for every instance of the circuit, such that $y_i^l=\smash{\hat{f}_i^l(\sum_j w_{ij}y_j^{l-1})}$, where $\smash{\hat{f}_i^l}$ approximately corresponds to $f$, but now depends on $i$ and $l$, and thus is different for every neuron.
 Note that mismatch in the weight parameters can be incorporated in a similar way. However, for illustrative purposes, and because in our implementation most of the effects can be modeled as perturbations of $f$, we focus on the effective heterogeneity of the nonlinearity here.

\subsection{Training with heterogeneous transfer functions}
\label{sect:training}

We consider the scenario where a loss function $\mathcal{L}(y^L, \hat{y})$ is optimized with respect to the model parameters $\Theta$ over a given training dataset through stochastic gradient descent. Here, $y^L$ is the output of the final layer, and $\hat{y}$ is the target signal.
Notably, the model with the heterogeneous activation functions can be optimized in the same way as a homogeneous model with $f_i^l = f \forall i,l$. All we need is an accurate enough, differentiable model of the actual physical system. In this way, the training becomes device-dependent. The resulting function, however, will be roughly equivalent to the homogeneous system (assuming both models are of similar capacity.)

The process of implementing a target functionality in such a heterogeneous system is illustrated in Fig.~\ref{fig:cartoon_method}.
Once a neural network architecture with modifiable weights is implemented in silicon, the response characteristics of the different circuit instances can be measured by controlling the inputs specific cells receive and recording their output at the same time (see suppl.~\ref{sect:si:measurements} for details).
The continuous, parameterized description is then used by the training algorithm, which is run on traditional computing hardware, to generate a network configuration that is tailored to the particular task and the physical device that has been characterized.

\subsection{Reducing circuits to their essence}
\label{sect:closedloop}
 
\begin{figure}[tp]
	\centering
	\includegraphics[scale=.7, trim=2cm 0 2cm 0]{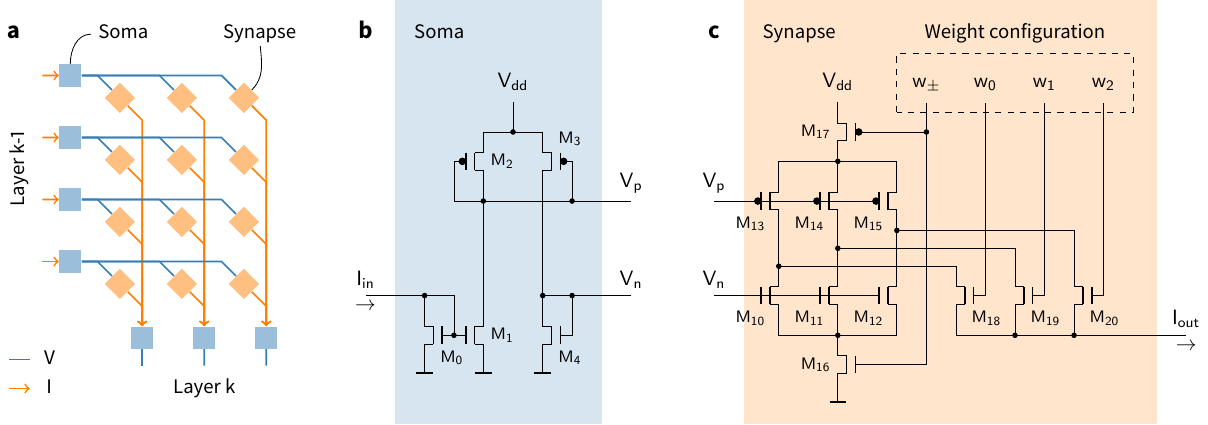}
	\caption{
A multi-layer neural network implemented with current-mode analog circuits.
 a) A network is constructed by connecting layers of soma circuits through matrices of synapse circuits. The output of a soma circuit is communicated as a voltage (blue) and passed to a row of synapse circuits, implementing multiplications by scalars.
 The output of a synapse is a current (orange), such that the outputs of a column of synapses can be summed up by simply connecting them through wires.
 The summed current is then passed as input to a soma of the next layer, which implements the nonlinearity.
 b) Proposed soma circuit, taking a current as input and providing two output voltages $V_n$ and $V_p$, which in the subthreshold region are proportional to the log-transformed, rectified input current. c) Proposed programmable synapse circuit with 3 bit precision, taking voltages $V_n$ and $V_p$ as inputs and providing an output current corresponding to an amplified version of the rectified soma input current, where the gain is set by the digital signals $w_\pm$, $w_i$.
	}
  \label{fig:circuits}
\end{figure}

Modeling the low-level properties of the physical substrate means that stabilization mechanisms and abstractions, such as digital logic, are not required to implement a given function.
However, this is feasible only if an accurate enough model of the physical system can be obtained sufficiently easily. At the core of our approach are therefore circuits which are both more compact than previously explored ones, and at the same time can be accurately described by a simple model as the one outlined above, where multiplication remains linear and mismatch effects can be modeled as deviations in $f$.
The particular circuits we propose are shown in Fig.~\ref{fig:circuits}.

\paragraph{Circuit description.}
 As illustrated in Fig.~\ref{fig:circuits}a, we implement a multilayer perceptron architecture by connecting multiple layers of ``soma'' circuits through matrices of ``synapse'' circuits.
 A soma circuit (Fig.~\ref{fig:circuits}b) takes a current as input and communicates its output in terms of voltages, which are passed as input signals to a row of synapse circuits.
 A synapse circuit (Fig.~\ref{fig:circuits}c), in turn, provides a current as output, such that the outputs of a column of synapses can be summed up simply by connecting them together.
 The resulting current is then fed as an input current to the somata of the next layer.
 The first transistor of the soma circuit rectifies the input current.
 The remaining elements of the soma circuit, together with a connected synapse circuit, form a set of scaling current mirrors, i.e. rudimentary amplifiers, a subset of which can be switched on or off to achieve a particular weight value by setting the respective synapse configuration bits.
 Thus, the output of a synapse corresponds to a scaled version of the rectified input current of the soma, similar to the ReLU transfer function.
 To achieve stable weight representations, we use digital memory elements to store the weight values.
 In our example implementation we use signed 3-bit synapses, which are based on $2 \times 3$ current mirrors of different dimensions (3 for positive and 3 for negative values).
 One of $2^4$ possible weight values is then selected by switching the respective current mirrors on or off.
 The scaling factor of a particular current mirror, and thus its contribution to the total weight value, is proportional to the ratio of the widths of the two transistors forming it.
 The weight configuration of an individual synapse are stored in memory elements that are part of the actual synapse circuit.
 Thus, in contrast to digital processing systems, our circuit computes \emph{in memory} and thereby avoids the bottleneck of expensive data transfer between memory and processing elements.
 A more detailed description of the circuits can be found in suppl.~\ref{sect:si:circuits}.

 The simple circuits presented here offer several advantages besides the fact that they can be implemented in small areas:
 First, numeric values are conveyed only through current mirrors, and therefore are temperature-independent.
 Second, most of the fabrication-induced variability is due to the devices in the soma with five consecutive transistors, whereas only one layer of transistors affects the signal in the synapse.
 This means that the synapse-induced mismatch can be neglected in a first order approximation.
 
 \paragraph{Subthreshold operation.}
 For reduced power dissipation, our circuits can be operated in the subthreshold regime.
 The subthreshold current of a transistor is exponential in the gate voltage, rather than polynomial as is the case for above threshold operation, and can span many orders of magnitude.
 Thus, a system based on this technology can be operated at orders of magnitude lower currents than a digital one.
 In turn, this means that the device mismatch arising due to imperfections in the fabrication process can have an exponentially larger impact.
 Fortunately, as our method neither depends on the specific form nor the magnitude of the mismatch, this does not pose an obstacle.

\paragraph{Device characterization.}

 Once an analog electronic neural network has been implemented physically as a VLSI device, the response characteristics of the individual circuits are obtained through measurements.
 The transfer function implemented by our circuits can be well described by a rectified linear curve, where the only free parameter is the slope, and thus can be determined from a single measurement per neuron.
 Specifically, the transfer curves of all neurons in a layer $l$ can be measured through a simple procedure:
 A single neuron in layer $l-1$ is connected, potentially through some intermediate neurons, to the input layer and is defined to be the `source'.
 Similarly, a neuron in layer $l+1$ is connected, potentially through intermediate neurons, to the output layer and is called the `monitor'.
 All neurons of layer $l$ can now be probed individually using the source and monitor neurons, whereby the signal to the input layer is held fixed and the signal recorded at the output layer is proportional to the slope of the measured neuron.
 Note that the absolute scale of the responses is not relevant, i.e. only the relative scale within one layer matters, as the output of individual layers can be scaled arbitrarily without altering the network function.
 The same procedure can be applied to all layers to obtain a complete characterization of the network.
 The measurements can be parallelized by defining multiple source and monitor neurons per measurement to probe several neurons in one layer simultaneously, or by introducing additional readout circuitry between layers to measure multiple layers simultaneously (see details in suppl.~\ref{sect:si:measurements}).

\section{Experimental results}

\begin{figure}[htb]
	\centering
	\includegraphics[scale=0.68]{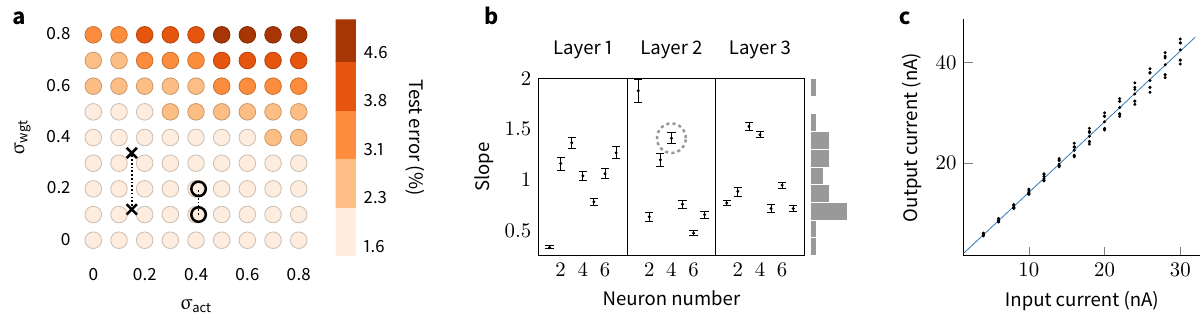}
	\caption{
    Robustness of the proposed method (left) and measured device characteristics (middle and right).
    a) Each point corresponds to the test error of a 196-200-200-10 network on \textsc{mnist} (each point averaged over 10 trained models).
    The variations in the activations (soma circuits) are fixed before training and the resulting slopes are used by the training algorithm.
    The mismatch in the synapses is not modeled explicitly, and is fixed only after training, during testing.
    The worst-case scenario (greatest observed mismatch) is marked by crosses for our simulated circuit, and by circles for our fabricated prototype chip for the smallest and largest weight value, respectively.
    b) Measured slopes for different neurons of the fabricated prototype device.
    c) Measurements corresponding to the circled point in (b) and line fitted to determine the slope. Note the strongly linear behavior of the circuit.
	}
  \label{fig:sweep_std}
\end{figure}

We validate our approach both through detailed simulations and an actual prototype chip that has been fabricated.
Large-scale circuit-level \textsc{spice} simulations of systems consisting of hundreds of thousands of transistors were performed to assess power consumption, processing speed, and the accuracy of such an analog implementation.
We further demonstrate the effectiveness of the whole pipeline by implementing a classification model on a prototype device, where training depends on measured device characteristics.
All models were trained using Adam \cite{kingma2014adam} and dual-copy rounding \cite{courbariaux2014low} to obtain signed 3-bit weight representations (see suppl.~\ref{sect:si:training} for training details.)

\subsection{Precise computation on imprecise devices}

For our current mirror-based circuits, the deviations resulting from mismatch in all the transistors is to a high level of accuracy modeled as deviations in the slopes of the activation functions $\smash{\hat{f}_i}$, where the distribution of the slopes corresponds to a log-normal distribution (see suppl.~\ref{sect:si:circuits} for details.) Similarly, the deviations of the weight values from the `ideal' values follow a log-normal distribution.
We evaluate the trainability of such a system through simulations, where we take into account the heterogeneous activations during training, and add noise to the weights at test time to simulate a scenario where we only measure the slopes of the activations, but not the exact weight deviations (see suppl.~\ref{sect:si:simulation} simulation for details).
As shown in fig.~\ref{fig:sweep_std}, an accurate model can be trained for a wide variety of mismatch conditions 
The level of mismatch is characterized by the standard deviation of the underlying normal distribution, $\smash{\sigma_\textrm{act}}$, and $\smash{\sigma_\textrm{wgt}}$, respectively.
Realistic mismatch conditions as encountered in our prototype device and in low-level circuit simulations are marked by crosses and circles, respectively.
Fig.~\ref{fig:sweep_std}b shows the slopes found in an actual device for different neurons. Fig.~\ref{fig:sweep_std}c shows all measurements taken to extract the slope of a single neuron.

\subsection{Handwritten and spoken digit classification}

\begin{figure}[ht]
	\centering
	\includegraphics[scale=0.8]{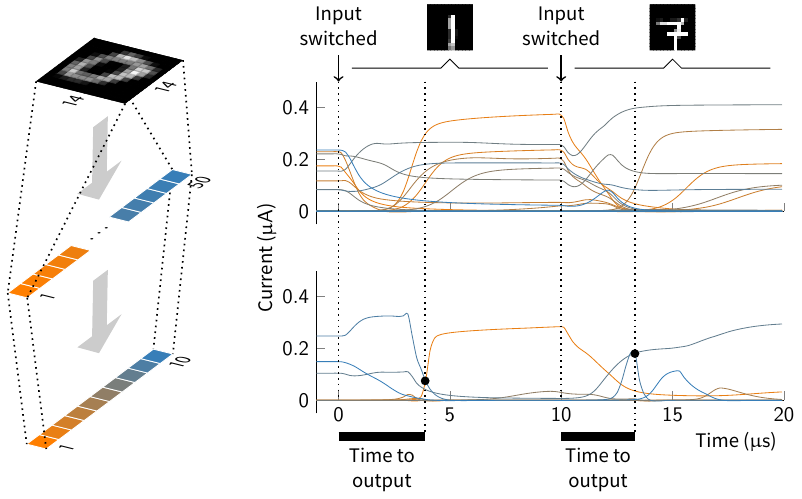}
	\caption{
 Analog circuit dynamics allow classification within microseconds. The curves represent the activities (currents) of all hidden (top) and output (bottom) units of a 196-50-10 network, as shown on the left. When a new input symbol is presented (top), the circuit converges to its new state within microseconds.
 Only a few units remain active, while many tend to zero, such that their soma circuits and connected synapses dissipate very little power.
 }
  \label{fig:dynamics}
\end{figure}

\begin{figure}[ht]
	\centering
	\includegraphics[scale=.69, trim=1cm 0 1cm 0]{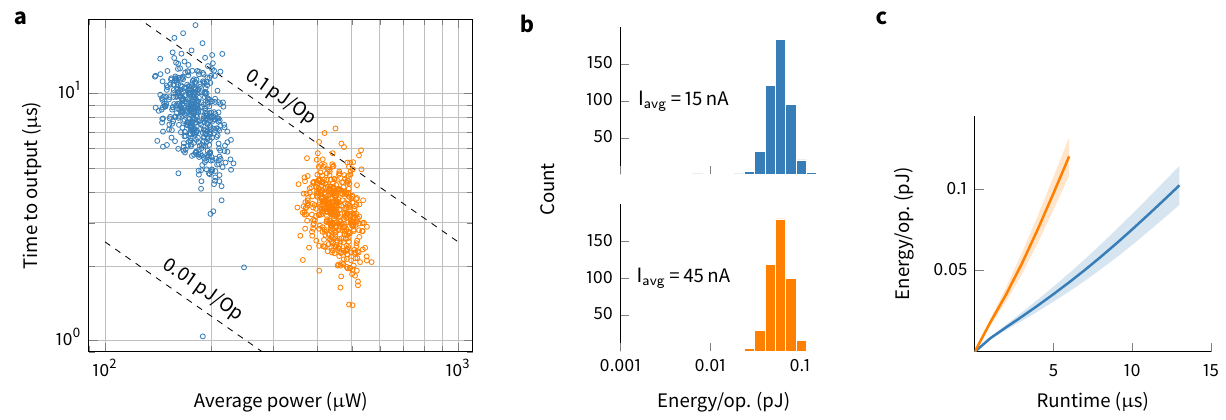}
	\caption{
 Processing performance of a network for handwritten digit classification. All data shown was generated by presenting 500 different input patterns from the \textsc{mnist} test set to a trained 196-100-50-10 network with the average input current per input neuron set to 15\,nA (blue) or 45\,nA (orange), respectively.
 a) The time to output is plotted  against the average power dissipated over the duration of the transient (from start of the input pattern until time to output).
 Changing the input current causes a shift along the equi-efficiency lines, that is, the network can be run slower or faster at the same efficiency (energy per operation).
 b) The average energy required per operation was computed from the data shown in (a). The data corresponds to the hypothetical ideal case were the network would be stopped as soon as the correct output is reached.
 c) Energy dissipated per operation for different run times, corresponding to different fixed rates at which inputs are presented (solid lines correspond to the mean; shaded areas to the standard deviation.)
 }
  \label{fig:performance}
\end{figure}

 We evaluate properties of the circuits, such as processing speed, power consumption and accuracy through low-level \textsc{spice} simulations.
 For this purpose we trained small 196-100-50-10 networks on the (downsampled) \textsc{mnist} dataset \cite{lecun1998mnist} and the \textsc{tidigits} dataset of spoken digits \cite{leonard1993tidigits}.
 By evaluating the responses of the simulated circuit on subsets of the respective test sets, its classification accuracy was found to be comparable to the abstract software neural network (see Tab.~\ref{tab:accuracy} for comparison).
 Fig.~\ref{fig:dynamics} shows how inputs are processed by a small example circuit implementing a 196-50-10 network, containing around 10k synapses and over 100k transistors.
 Starting with the presentation of an input pattern in the top layer, where currents are proportional to input stimulus intensity, the higher layers react almost instantaneously and provide the correct classification, i.e. the index of the maximally active output unit, within a few microseconds.
 After a switch of input patterns, the signals quickly propagate through the network and the outputs of different nodes converge to their asymptotic values.
 The time it takes the circuit to converge to its final output defines the `time to output', constraining the maximum frequency at which input patterns can be presented and evaluated correctly.
 Measured convergence times are summarized in Fig.~\ref{fig:performance} for different patterns from the \textsc{mnist} test set, and are found to be in the range of microseconds for a trained 196-100-50-10 network, containing over 25k synapses and around 280k transistors.
 Note that the observed timescale is not fixed, as the network can be run faster or slower by changing the input current, while the average energy dissipated per operation remains roughly constant (fig.~\ref{fig:performance}b).

 The processing efficiency of the system (energy per operation) was computed for different input patterns by integrating the power dissipated between the time at which the input pattern was switched and the time to output.
 Fig.~\ref{fig:performance} shows the processing efficiency for the same network with different input examples and under different operating currents (see details on the performance measurements in suppl.~\ref{sect:si:performance}).
 With the average input currents scaled to either 15 or 45\,nA per neuron respectively, the network takes several microseconds to converge and consumes tens or hundreds of microwatts in total, which amounts to a few nanowatts per multiply-accumulate operation.
 With the supply voltage set to 1.8\,V, this corresponds to less than 0.1\,pJ per operation in most cases.
 With the average input current set to 15\,nA per neuron, the network produces the correct output within 15\,$\upmu$s in over 99\,\% of all cases (mean 8.5\,$\upmu$s; std. 2.3\,$\upmu$s).
 Running the circuit for 15\,$\upmu$s requires $0.12\pm0.01$\,pJ per operation, such that about 1.7 trillion multiply-accumulate operations can be computed per second at a power budget of around 200\,$\upmu$W if input patterns are presented at a rate of 66\,kHz.
 Without major optimizations to either process or implementation, this leads to an efficiency of around 8\,TOp/J, to our knowledge a performance greater than that achieved by digital single-purpose neural network accelerators in similar scenarios \cite{cavigelli2015origami,park2016energy}.
 General purpose digital systems are far behind such specialized systems in terms of efficiency, with the Volta GPU generation achieving a maximum of 0.05 TOp/J (float32) or 0.4 TOp/J (tensor core) \cite{nvidia2017}.

  Tab.~\ref{tab:accuracy} summarizes the classification accuracy for different architectures and datasets for a software simulation of an ideal network without mismatch, a behavioral simulation of the heterogeneous system, and the full circuit simulation of the hardware device. The computed power efficiency is shown for the different architectures.

 \begin{table}[tb]
 \centering
 \begin{tabular}{lll}
	 & $14\times14$ \textsc{mnist} & \textsc{tidigits} \\
	 \hline
	 {Homogeneous model mean / best accuracy (\%)} & $97.6\pm0.1$ / 98.0 & $87.3\pm4.2$ / 93.4 \\
	 {Inhomogeneous model mean / best accuracy (\%)} &  $97.6\pm0.2$ / 98.0  &  $88.0\pm3.8$ / 94.3 \\
	 {\textsc{spice} simulation accuracy (\%)} & 98.0 & 94.6 \\
	 {Energy-efficiency (TOp/J)} & 7.97 & 6.39 \\
	 \hline
 \end{tabular}
 \caption{
     Classification accuracy and power-efficiency of a 196-100-50-10 network trained on the \textsc{mnist} and \textsc{tidigits} datasets. Behavioral results are averaged over 10 models with different initializations. The parameters of the best performing one out of the 10 models were used in the \textsc{spice} circuit simulations. As detailed circuit simulations are computationally expensive, subsets of the actual test sets were used to compute the classification accuracy of the simulated circuits (the first 500 samples from the \textsc{mnist} test set; 500 random samples from the \textsc{tidigits} test set).
 }
 \label{tab:accuracy}
 \end{table}

 \subsection{VLSI implementation}

 As a closed-loop demonstration of our framework, we designed a prototype VLSI chip and trained it for a classification task.
 A design based on the circuits shown in Fig.~\ref{fig:circuits}, containing three layers of seven neurons each, was fabricated in 180\,nm CMOS technology (see details in suppl.~\ref{sect:si:prototype}).
 After characterizing the individual neuron circuits through measurements as described in sect.~\ref{sect:closedloop} we trained a 4-7-3 network on 80\,\% of the Iris flower dataset \cite{iris}, programmed the device with the found parameters, and used the remaining 20\,\% of the data to test the classification performance.
 The hardware implementation was able to classify 100\% of the test data correctly.
 Measured device characteristics are shown in fig.~\ref{fig:sweep_std}.

 \section{Discussion}

The simplicity of analog VLSI circuits implementing elementary operations means that a massively parallel system can be fully laid out in hardware, without the need for time-multiplexed processing elements. This, in turn, allows for memory and processing elements to be co-located, eliminating the data transfer bottleneck between them.
 Using digital technology, such fully parallel implementations would quickly become prohibitively large due to the much greater circuit complexity of digital processing elements. While the focus in this work has been on efficient analog VLSI implementations, an equivalent approach can be used with other substrates, such as memristive processors, which similarly suffer from fabrication-induced mismatch \cite{ambrogio2013spike,kim2011functional,prezioso2015training,niu2010impact}.
 In fact, any system that can be properly characterized and has configurable elements stands to benefit from this approach.
 
While, as a proof-of-concept, we evaluated the approach on multilayer perceptron architectures, other structures, such as convnets, and even recurrent architectures, such as LSTMs \cite{hochreiter1997long} can be trained in an analogous way.
However, if all weights are to be implemented explicitly in silicon, the system design here, while potentially very fast, would not necessarily benefit from the small memory footprint achieved via weight sharing in traditional convnet implementations.

In the current setting, the efficiency of our system is limited by the worst-case per-example runtime, i.e. there may be a few examples where outputs require significantly longer to converge to the correct classification result than the majority.
This can lead to unnecessarily long presentation times for many examples, thereby causing unnecessary power consumption. Smart methods for estimating presentation times from the input data could accelerate convergence for slowly converging samples by using higher input currents, and conversely, faster samples could be slowed down to lower the variability of convergence times and overall reduce energy consumption.
Future research will focus on such estimators, and alternatively explore ways of reducing convergence time variability during network training.

This proof-of-principle study is a step towards large scale, possibly ultra-low-power analog VLSI deep neural network processors, paving the way for specialized applications which had been infeasible before due to speed or power constraints. Small, efficient implementations could enable autonomous systems to achieve almost immediate reaction times under strict power limitations. Scaled-up versions can allow for substantially more efficient processing in data centers, a greatly reduced energy footprint, or permitting substantially more data to be effectively processed.

\paragraph{Acknowledgments:} We are very grateful to Ning Qiao for his help with the VLSI implementation. The work was supported by the Swiss National Science Foundation grant SpikeComp (200021\_146608) and the University of Zurich.
	
\paragraph{Author contributions:} The hardware architectures and circuit simulations were developed by J.B.; the VLSI prototype was tested by J.B.; the training software was developed by J.B. and D.N.; the \textsc{tidigits} data was pre-processed by D.N.; the experiments were carried out and the data was analyzed by J.B.; the manuscript was written by J.B., D.N., G.I., S.L., and M.P.

\bibliographystyle{plain}
\bibliography{bibliography}

\begin{thebibliography}{10}

\bibitem{Alspector_Allen87}
J.~Alspector and R.B. Allen.
\newblock A neuromorphic {VLSI} learning system.
\newblock In P.~Losleben, editor, {\em {P}roceedings of the 1987 Stanford
  Conference on Advanced Research in {VLSI}}, pages 313--349, Cambridge, MA,
  USA, 1987. MIT Press.

\bibitem{ambrogio2013spike}
S~Ambrogio, S~Balatti, F~Nardi, S~Facchinetti, and D~Ielmini.
\newblock Spike-timing dependent plasticity in a transistor-selected resistive
  switching memory.
\newblock {\em Nanotechnology}, 24(38):384012, 2013.

\bibitem{andreou1991current}
Andreas~G Andreou, Kwabena Boahen, Philippe~O Pouliquen, Aleksandra Pavasovic,
  Robert~E Jenkins, Kim Strohbehn, et~al.
\newblock Current-mode subthreshold {MOS} circuits for analog {VLSI} neural
  systems.
\newblock {\em {IEEE} Transactions on neural networks}, 2(2):205--213, 1991.

\bibitem{backus1978}
John Backus.
\newblock Can programming be liberated from the {von Neumann} style?: A
  functional style and its algebra of programs.
\newblock {\em Commun. ACM}, 21(8):613--641, 1978.

\bibitem{Borgstrom_etal90}
T.H. Borgstrom, M~Ismail, and S.B. Bibyk.
\newblock Programmable current-mode neural network for implementation in
  analogue {MOS} {VLSI}.
\newblock {\em IEE Proceedings G}, 137(2):175--184, 1990.

\bibitem{cavigelli2015origami}
Lukas Cavigelli, David Gschwend, Christoph Mayer, Samuel Willi, Beat Muheim,
  and Luca Benini.
\newblock Origami: A convolutional network accelerator.
\newblock In {\em Proceedings of the 25th edition on Great Lakes Symposium on
  VLSI}, pages 199--204. ACM, 2015.

\bibitem{chen2016elm}
Y.~Chen, E.~Yao, and A.~Basu.
\newblock A 128-channel extreme learning machine-based neural decoder for brain
  machine interfaces.
\newblock {\em IEEE Transactions on Biomedical Circuits and Systems},
  10(3):679--692, 2016.

\bibitem{chen201614}
Yu-Hsin Chen, Tushar Krishna, Joel Emer, and Vivienne Sze.
\newblock 14.5 eyeriss: An energy-efficient reconfigurable accelerator for deep
  convolutional neural networks.
\newblock In {\em 2016 {IEEE} International Solid-State Circuits Conference
  ({ISSCC})}, pages 262--263. IEEE, 2016.

\bibitem{chen2014dadiannao}
Yunji Chen, Tao Luo, Shaoli Liu, Shijin Zhang, Liqiang He, Jia Wang, Ling Li,
  Tianshi Chen, Zhiwei Xu, Ninghui Sun, et~al.
\newblock Dadiannao: A machine-learning supercomputer.
\newblock In {\em Microarchitecture, 2014 47th Annual {IEEE}/{ACM}
  International Symposium on}, pages 609--622. IEEE, 2014.

\bibitem{choudhary2012silicon}
Swadesh Choudhary, Steven Sloan, Sam Fok, Alexander Neckar, Eric Trautmann,
  Peiran Gao, Terry Stewart, Chris Eliasmith, and Kwabena Boahen.
\newblock Silicon neurons that compute.
\newblock In {\em International conference on artificial neural networks},
  pages 121--128. Springer, 2012.

\bibitem{courbariaux2014low}
Matthieu Courbariaux, Yoshua Bengio, and Jean-Pierre David.
\newblock Low precision arithmetic for deep learning.
\newblock {\em arXiv preprint arXiv:1412.7024}, 2014.

\bibitem{courbariaux2015binaryconnect}
Matthieu Courbariaux, Yoshua Bengio, and Jean-Pierre David.
\newblock Binaryconnect: Training deep neural networks with binary weights
  during propagations.
\newblock In {\em Advances in Neural Information Processing Systems}, pages
  3105--3113, 2015.

\bibitem{delbruck201032}
Tobi Delbruck, Raphael Berner, Patrick Lichtsteiner, and Carlos Dualibe.
\newblock 32-bit configurable bias current generator with sub-off-current
  capability.
\newblock In {\em Proceedings of 2010 {IEEE} International Symposium on
  Circuits and Systems}, pages 1647--1650. IEEE, 2010.

\bibitem{eliasmith2004neural}
C.~Eliasmith and C.H. Anderson.
\newblock {\em Neural Engineering: Computation, Representation, and Dynamics in
  Neurobiological Systems}.
\newblock A Bradford book. MIT Press, 2004.

\bibitem{farabet2012comparison}
Cl{\'e}ment Farabet, R~Paz-Vicente, JA~P{\'e}rez-Carrasco, Carlos
  Zamarre{\~n}o-Ramos, Alejandro Linares-Barranco, Yann LeCun, Eugenio
  Culurciello, Teresa Serrano-Gotarredona, and Bernabe Linares-Barranco.
\newblock Comparison between frame-constrained fix-pixel-value and frame-free
  spiking-dynamic-pixel convnets for visual processing.
\newblock {\em Frontiers in Neuroscience}, 6:1--12, 2012.

\bibitem{iris}
RA~Fisher.
\newblock Iris flower data set, 1936.

\bibitem{gokhale2014240}
Vinayak Gokhale, Jonghoon Jin, Aysegul Dundar, Ben Martini, and Eugenio
  Culurciello.
\newblock A 240 g-ops/s mobile coprocessor for deep neural networks.
\newblock In {\em Computer Vision and Pattern Recognition Workshops ({CVPRW}),
  2014 {IEEE} Conference on}, pages 696--701. IEEE, 2014.

\bibitem{griffin199111}
Matthew Griffin, Gary Tahara, Kurt Knorpp, Ray Pinkham, and Bob Riley.
\newblock An 11-million transistor neural network execution engine.
\newblock In {\em Solid-State Circuits Conference, 1991. Digest of Technical
  Papers. 38th ISSCC., 1991 {IEEE} International}, pages 180--313. IEEE, 1991.

\bibitem{han2015deep}
Song Han, Huizi Mao, and William~J Dally.
\newblock Deep compression: Compressing deep neural networks with pruning,
  trained quantization and huffman coding.
\newblock {\em arXiv preprint arXiv:1510.00149}, 2015.

\bibitem{hasler2013finding}
Jennifer Hasler and Bo~Marr.
\newblock Finding a roadmap to achieve large neuromorphic hardware systems.
\newblock {\em Frontiers in neuroscience}, 7, 2013.

\bibitem{hay1960mark}
John~C Hay, Ben~E Lynch, and David~R Smith.
\newblock Mark i perceptron operators' manual.
\newblock Technical report, CORNELL AERONAUTICAL LAB INC BUFFALO NY, 1960.

\bibitem{hinton2015distilling}
Geoffrey Hinton, Oriol Vinyals, and Jeff Dean.
\newblock Distilling the knowledge in a neural network.
\newblock {\em arXiv preprint arXiv:1503.02531}, 2015.

\bibitem{hochreiter1997long}
Sepp Hochreiter and J{\"u}rgen Schmidhuber.
\newblock Long short-term memory.
\newblock {\em Neural computation}, 9(8):1735--1780, 1997.

\bibitem{indiveri2015neuromorphic}
Giacomo Indiveri, Federico Corradi, and Ning Qiao.
\newblock Neuromorphic architectures for spiking deep neural networks.
\newblock In {\em {IEEE} International Electron Devices Meeting ({IEDM})},
  2015.

\bibitem{indiveri2015memory}
Giacomo Indiveri and Shih-Chii Liu.
\newblock Memory and information processing in neuromorphic systems.
\newblock {\em Proceedings of the {IEEE}}, 103(8):1379--1397, 2015.

\bibitem{kim2011functional}
Kuk-Hwan Kim, Siddharth Gaba, Dana Wheeler, Jose~M Cruz-Albrecht, Tahir
  Hussain, Narayan Srinivasa, and Wei Lu.
\newblock A functional hybrid memristor crossbar-array/{CMOS} system for data
  storage and neuromorphic applications.
\newblock {\em Nano letters}, 12(1):389--395, 2011.

\bibitem{kingma2014adam}
Diederik Kingma and Jimmy Ba.
\newblock Adam: A method for stochastic optimization.
\newblock {\em arXiv preprint arXiv:1412.6980}, 2014.

\bibitem{Lakshmikumar_1986}
Kadaba~R Lakshmikumar, Robert Hadaway, and Miles Copeland.
\newblock Characterisation and modeling of mismatch in {MOS} transistors for
  precision analog design.
\newblock {\em {IEEE} Journal of Solid-State Circuits}, 21(6):1057--1066, 1986.

\bibitem{lecun2015deep}
Yann LeCun, Yoshua Bengio, and Geoffrey Hinton.
\newblock Deep learning.
\newblock {\em Nature}, 521(7553):436--444, 2015.

\bibitem{lecun1998mnist}
Yann LeCun, Corinna Cortes, and Christopher~JC Burges.
\newblock The {MNIST} database of handwritten digits, 1998.

\bibitem{leonard1993tidigits}
R~Gary Leonard and George Doddington.
\newblock Tidigits speech corpus.
\newblock {\em Texas Instruments, Inc}, 1993.

\bibitem{masa1994high}
Peter Masa, Klaas Hoen, and Hans Wallinga.
\newblock A high-speed analog neural processor.
\newblock {\em Micro, {IEEE}}, 14(3):40--50, 1994.

\bibitem{merolla668}
Paul~A Merolla, John~V Arthur, Rodrigo Alvarez-Icaza, Andrew~S Cassidy, Jun
  Sawada, Filipp Akopyan, Bryan~L Jackson, Nabil Imam, Chen Guo, Yutaka
  Nakamura, Bernard Brezzo, Ivan Vo, Steven~K Esser, Rathinakumar Appuswamy,
  Brian Taba, Arnon Amir, Myron~D Flickner, William~P Risk, Rajit Manohar, and
  Dharmendra~S Modha.
\newblock {A million spiking-neuron integrated circuit with a scalable
  communication network and interface}.
\newblock {\em Science}, 345(6197):668--673, 2014.

\bibitem{moons2018binareye}
Bert Moons, Daniel Bankman, Lita Yang, Boris Murmann, and Marian Verhelst.
\newblock Binareye: An always-on energy-accuracy-scalable binary cnn processor
  with all memory on chip in 28nm cmos.
\newblock In {\em 2018 IEEE Custom Integrated Circuits Conference (CICC)},
  pages 1--4. IEEE, 2018.

\bibitem{mundy2015efficient}
Andrew Mundy, James Knight, Terrence~C Stewart, and Steve Furber.
\newblock An efficient spinnaker implementation of the neural engineering
  framework.
\newblock In {\em 2015 International Joint Conference on Neural Networks
  (IJCNN)}, pages 1--8. IEEE, 2015.

\bibitem{niu2010impact}
Dimin Niu, Yiran Chen, Cong Xu, and Yuan Xie.
\newblock Impact of process variations on emerging memristor.
\newblock In {\em Design Automation Conference ({DAC}), 2010 47th {ACM/IEEE}},
  pages 877--882. IEEE, 2010.

\bibitem{nvidia2017}
NVIDIA.
\newblock Nvidia tesla v100 gpu architecture. wp-08608-001.
\newblock {\em NVIDIA Whitepaper}, 2017.

\bibitem{oconnor2013real}
Peter O'Connor, Daniel Neil, Shih-Chii Liu, Tobi Delbruck, and Michael
  Pfeiffer.
\newblock Real-time classification and sensor fusion with a spiking deep belief
  network.
\newblock {\em Frontiers in Neuromorphic Engineering}, 7, 2013.

\bibitem{park2016energy}
SW~Park, J~Park, K~Bong, D~Shin, J~Lee, S~Choi, and HJ~Yoo.
\newblock An energy-efficient and scalable deep learning/inference processor
  with tetra-parallel {MIMD} architecture for big data applications.
\newblock {\em {IEEE} transactions on biomedical circuits and systems}, 2016.

\bibitem{pei2019towards}
Jing Pei, Lei Deng, Sen Song, Mingguo Zhao, Youhui Zhang, Shuang Wu, Guanrui
  Wang, Zhe Zou, Zhenzhi Wu, Wei He, et~al.
\newblock Towards artificial general intelligence with hybrid tianjic chip
  architecture.
\newblock {\em Nature}, 572(7767):106--111, 2019.

\bibitem{Pelgrom_1998}
Marcel~JM Pelgrom, Hans~P Tuinhout, and Maarten Vertregt.
\newblock Transistor matching in analog {CMOS} applications.
\newblock {\em IEDM Tech. Dig}, pages 915--918, 1998.

\bibitem{Pelgrom_1989}
M.J.M. Pelgrom, Aad~C.J. Duinmaijer, and A.P.G. Welbers.
\newblock Matching properties of {MOS} transistors.
\newblock {\em {IEEE} Journal of Solid-State Circuits}, 24(5):1433--1439, 10
  1989.

\bibitem{prezioso2015training}
Mirko Prezioso, Farnood Merrikh-Bayat, BD~Hoskins, GC~Adam, Konstantin~K
  Likharev, and Dmitri~B Strukov.
\newblock Training and operation of an integrated neuromorphic network based on
  metal-oxide memristors.
\newblock {\em Nature}, 521(7550):61--64, 2015.

\bibitem{rosenblatt1958}
F.~Rosenblatt.
\newblock {The perceptron: a probabilistic model for information storage and
  organization in the brain.}
\newblock {\em Psychological review}, 65(6):386--408, 11 1958.

\bibitem{Satyanarayana_etal92}
S.~Satyanarayana, Y.P. Tsividis, and H.P. Graf.
\newblock A reconfigurable {VLSI} neural network.
\newblock {\em {IEEE} Journal of Solid-State Circuits}, 27(1):67--81, 1 1992.

\bibitem{scherer2010accelerating}
Dominik Scherer, Hannes Schulz, and Sven Behnke.
\newblock Accelerating large-scale convolutional neural networks with parallel
  graphics multiprocessors.
\newblock In {\em Artificial Neural Networks--ICANN 2010}, pages 82--91.
  Springer, 2010.

\bibitem{thakur2016elm}
C.~S. Thakur, R.~Wang, T.~J. Hamilton, J.~Tapson, and A.~van Schaik.
\newblock A low power trainable neuromorphic integrated circuit that is
  tolerant to device mismatch.
\newblock {\em IEEE Transactions on Circuits and Systems I: Regular Papers},
  63(2):211--221, 2 2016.

\bibitem{Vittoz90}
E.A. Vittoz.
\newblock Analog {VLSI} implementation of neural networks.
\newblock In {\em Proc. {IEEE} Int. Symp. Circuit and Systems}, pages
  2524--2527, New Orleans, 1990.

\bibitem{voelker2017extending}
Aaron~R Voelker, Ben~V Benjamin, Terrence~C Stewart, Kwabena Boahen, and Chris
  Eliasmith.
\newblock Extending the neural engineering framework for nonideal silicon
  synapses.
\newblock In {\em 2017 IEEE International Symposium on Circuits and Systems
  (ISCAS)}, pages 1--4. IEEE, 2017.

\end{thebibliography}

\pagebreak

\appendix
\section{Supplementary material}
    
 \subsection{Description of the example circuit}
 \label{sect:si:circuits}

 The example networks described in Sect.~\ref{sect:closedloop} have been implemented based on the circuits shown in Fig.~\ref{fig:circuits}.
 With $M_0$ as a diode-connected nFET, the soma circuit essentially performs a rectification of the input current $I_\mathrm{in}$.
 Further, the current is copied to $M_1$ and, through $M_2$ and $M_3$, also to $M_4$, such that $M_2$ together with pFETs from connected synapse circuits, as well as $M_4$ together with nFETs from connected synapse circuits form scaling current mirrors, generating scaled copies of the rectified input current $I_\mathrm{in}$.
 The scaling factor is thereby determined by the dimensions of $M_{10}$ to $M_{15}$.
 The transistors $M_{16}$ to $M_{20}$ operate as switches and are controlled by the digital signals $w_\pm$ $w_0$, $w_1$, and $w_2$.
 The value of $w_\pm$ determines whether the positive branch (pFETs $M_{13}$ to $M_{15}$; adding current to the node $I_\mathrm{out}$) or the negative branch (nFETs $M_{10}$ to $M_{12}$; subtracting current from the node $I_\mathrm{out}$) is switched on and thereby the sign of the synaptic multiplication factor.
 Setting $w_0$, $w_1$, and $w_2$ allows switching on or off specific contributions to the output current.
 In the example implementation the widths of $M_{10}$ to $M_{12}$, and $M_{13}$ to $M_{15}$, respectively, were scaled by powers of 2 (see Tab.~\ref{tab:trans_params}), such that a synapse would implement a multiplication by a factor approximately corresponding to the binary value of $(w_0, w_1, w_2)$.
 While our results are based on a signed 3-bit version of the circuit, arbitrary precision can be implemented by changing the number of scaling transistors and corresponding switches.
 The dimensions of $M_3$ and $M_4$ were adjusted such that the currents through transistors of the positive and the negative branch of one particular bit of a synapse were roughly matched when switched on.

 \begin{table}[htb]
 \centering
 \caption{Transistor dimensions used in all circuit simulations.}
 \bigskip
 \begin{tabular}{lccc}
	 Device		& W ($\upmu$m)	& L ($\upmu$m)	& W/L \\
	 \hline
	 $M_0 - M_4$		& 2.7		& 0.45		& 6	\\
	 $M_{10},~M_{13}$	& 0.27		& 0.54		& 0.5	\\
	 $M_{11},~M_{14}$	& 0.54		& 0.54		& 1	\\
	 $M_{12},~M_{15}$	& 1.08		& 0.54		& 2	\\
	 $M_{16} - M_{20}$	& 0.54		& 0.54		& 1     \\
	 \hline
 \end{tabular}
 \label{tab:trans_params}
 \end{table}

 Multilayer networks were constructed using the circuits described above by connecting layers of soma circuits through matrices made up of synapse circuits.
 The first stage of a network constructed in this way thereby is a layer of soma circuits, rather than a weight matrix, as is typically the case in artificial neural network implementations.
 This is because we prefer to provide input currents rather than voltages and only soma circuits take currents as inputs.
 As a consequence, due to the rectification, our network can not handle negative input signals.
 To obtain current outputs rather than voltages, one synapse is connected to each unit of the output layer and its weight set to 1 to convert the output voltages to currents.

 \subsection{Circuit simulation details}
 \label{sect:si:simulation}

 All circuits were simulated using \textsc{ngspice} release 26 and \textsc{bsim3} version 3.3.0 models of a \textsc{tsmc} 180\,nm process.
 The \textsc{spice} netlist for a particular network was generated using custom Python software and then passed to \textsc{ngspice} for DC and transient simulations.
 Input patterns were provided to the input layer by current sources fixed to the respective values.
 The parameters from Tab.~\ref{tab:trans_params} were used in all simulations and $V_\mathrm{dd}$ was set to 1.8\,V.
 Synapses were configured by setting their respective configuration bits $w_\pm$, $w_0$, $w_1$, and $w_2$ to either $V_\mathrm{dd}$ or ground, emulating a digital memory element.
 The parasitic capacitances and resistances to be found in an implementation of our circuits were estimated from post-layout simulations of single soma and synapse cells.
 The main slowdown of the circuit can be attributed to the parasitic capacitances of the synapses, which were found to amount to 11\,fF per synapse.

 Individual hardware instances of our system were simulated by randomly assigning small deviations to all transistors of the circuit.
 Since the exact nature of mismatch is not relevant for our main result (our training method compensates for any kind of deviation, regardless of its cause), the simple but common method of threshold matching was applied to introduce device-to-device deviations \cite{Lakshmikumar_1986}.
 Specifically, for every device, a shift in threshold voltage was drawn from a Gaussian distribution with zero mean and standard deviation $\sigma_{\Delta VT}=A_{VT}/\sqrt{W/L}$, where the proportionality constant $A_{VT}$ was set to 3.3\,mV$\upmu$m, approximately corresponding to measurements from a 180\,nm process \cite{Pelgrom_1998}.

 \subsection{Characterization of the simulated circuit}
 \label{sect:si:measurements}

\begin{figure}[tp]
	\centering
	\includegraphics[scale=0.9]{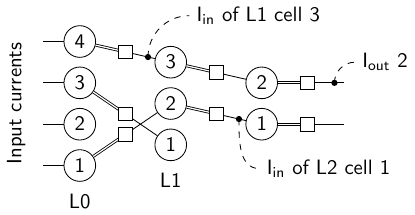}
	\caption{
 Illustration of the measurement procedure applied to the simulated circuits.
 The diagram shows one possible weight configuration that might come up during the parameter extraction procedure of a network with one input, one hidden, and one output layer.
 Circles represent soma circuits and squares synapse circuits. Voltages are represented by double lines, whereas currents are represented by single lines.
 Only synapses set to non-zero values are shown.
 Every unit receives exactly one input signal, and produces, together with a connected synapse circuit, at maximum one output current, which can be measured as the input to a unit of the consecutive layer.
 The input to the network is provided in terms of a set of input currents, the output is transformed to currents by means of an additional array of synapses after the last layer.
	}
	\label{fig:meas_illust}
\end{figure}

 To determine the transfer curves of individual neurons, the input-output relations of the respective soma circuits need to be measured.
 To save simulation time, a parallel measurement scheme was applied, based on the assumption that each neuron can be measured directly, rather than just the neurons in the output layer.
 Rather than measuring the log domain output voltages $V_n$ and $V_p$ we chose to record the input currents $I_\mathrm{in}$ to subsequent layers.
 The advantages of this approach are that quantities are not log-transformed and that potential distortions arising from the synapse circuits are taken into account.
 Furthermore, with this method only one probe is required per neuron, rather than two separate ones for in- and output signals.
 Moreover, the unit weight of a synapse (which is not know a priori) here becomes a property of the soma, so that weights are automatically normalized.
 To determine the transfer curves of the units in the different layers the weights were set to a number of different configurations and the input currents to the various units were measured for different input patterns provided to the network.
 Specifically, by setting the respective synapse circuits to their maximum value, every unit was configured to receive input from exactly one unit of the previous layer.
 One such configuration is shown in Fig.~\ref{fig:meas_illust}.
 The input currents to all units of the input layer were then set to the same value and the inputs to the units of the deeper layers were recorded.
 By generating many such connectivity patterns by permuting the connectivity matrix, and setting the input currents to different values, multiple data points (input-output relations) were recorded for each unit, such that continuous transfer curves could be fitted to the data.
 For the example networks described in Sect.~\ref{sect:closedloop}, 40 measurements turned out to be sufficient, resulting in roughly 10 data points per unit.
 Rectified linear functions $f(r) = \max\{0, a\cdot r\}$ were fitted to the data and the resulting parameters $a$ were used as part of the training algorithm.
 The parameters were normalized layer-wise to a mean slope of 1.
 Even though the sizes of the transistors implementing the positive and negative weight contributions are identical, their responses are not matched.
 To characterize their relative contributions, inputs were given to neurons through positive and negative connections simultaneously. Comparing the neuron response to its response with the negative connection switched off allows to infer the strength of the unit negative weight, which can then be used in the training algorithm.

\begin{figure}[tp]
	\centering
	\includegraphics[scale=0.9]{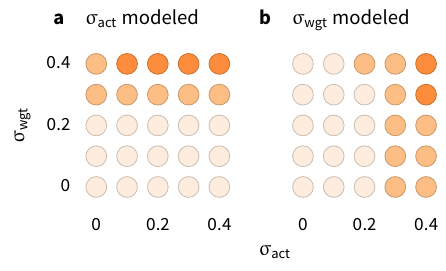}
	\caption{
	    Dependence of the network performance on the bit precision of the weights.
    Each point corresponds to the classification error of a $196-200-200-10$ network on the \textsc{mnist} test set (averaged over 10 trained networks).
	}
  \label{fig:sweep_bits}
\end{figure}

 \subsection{Training and evaluation details}
 \label{sect:si:training}

The $196-100-50-10$ networks were trained on the \textsc{mnist} and \textsc{tidigits} datasets using the \textsc{adam} optimizer \cite{kingma2014adam} and the mean squared error as loss function.
 The low-precision training (three signed bits per synapse) was done using a high-precision store and low-precision activations in the manner of the method described in \cite{courbariaux2014low}.
 An L1 regularization scheme was applied to negative weights only to reduce the number of negative inputs to neurons, as they would slow down the circuits.

 Different sets of empirically found hyperparameters were used during training for the \textsc{mnist} and \textsc{tidigits} datasets.
 A reduced resolution version ($14\times14$ pixels) of the \textsc{mnist} dataset was generated by identifying the 196 most active pixels (highest average value) in the dataset and only using those as input to the network.
 The single images were normalized to a mean pixel value of 0.04.
 The learning rate was set to 0.0065, the L1 penalty for negative weights was set to $10^{-6}$, and the networks were trained for 50 epochs with batch sizes of 200.

 Each spoken digit of the \textsc{tidigits} dataset was converted to 12 mel-spectrum cepstral coefficients (MFCCs) per time slice, with a maximum frequency of 8 kHz and a minimum frequency of 0 kHz, using 2048 FFT points and a skip duration of 1536 samples.  To convert the variable-length \textsc{tidigits} data to a fixed-size input, the input was padded to a maximum length of 11 time slices, forming a 12x11 input for each digit.  First derivative and second derivatives of the MFCCs were not used.  To increase robustness, a stretch factor was applied, changing the skip duration of the MFCCs by a factor of 0.8, 0.9, 1.0, 1.1, and 1.3, allowing fewer or more columns of data per example, as this was found to increase accuracy and model robustness.  A selection of hyperparameters for the MFCCs were evaluated, with these as the most successful.
 The resulting dataset was scaled pixel-wise to values between 0 and 1. Individual samples were then scaled to yield a mean value of 0.03.
 The networks were trained for 512 epochs on batches of size 200 with the learning rate set to 0.0073, and the L1 penalty to $10^{-6}$.


 \subsection{Performance measurements}
 \label{sect:si:performance}

 The accuracy of the abstract software model was determined after training by running the respective test sets through the network.
 Due to prohibitively long simulation times, only subsets of the respective test sets were used to determine the accuracy of the \textsc{spice}-simulated circuits. Specifically, the first 500 samples of the \textsc{mnist} test set and 500 randomly picked samples from the \textsc{tidigits} test set were used to obtain an estimate of the classification accuracy of the simulated circuits.
 The data was presented to the networks in terms of currents, by connecting current sources to the $I_\mathrm{in}$ nodes of the input layer.
 Individual samples were scaled to yield mean input currents of 15\,nA or 45\,nA per pixel, respectively.
 The time to output for a particular pattern was computed by applying one (random) input pattern from the test set and then, once the circuit had converged to a steady state, replaced by the input pattern to be tested.
 In this way, the more realistic scenario of a transition between two patterns is simulated, rather than a `switching on' of the circuit.
 The transient analysis was run for 7\,$\upmu$s and 15\,$\upmu$s with the mean input strength set to 45\,nA and 15\,nA, respectively, and a maximum step size of 20\,ns.
 At any point in time, the output class of the network was defined as the index of the output layer unit that was the most active.
 The time to output for each pair of input patterns was determined by checking at which time the output class of the network corresponded to its asymptotic state (determined through an operating point analysis of the circuit with the input pattern applied) and would not change anymore.
 The energy consumed by the network in a period of time was computed by integrating the current dissipated by the circuit over the decision time and multiplying it by the value of $V_\mathrm{dd}$ (1.8\,V in all simulations).

 \subsection{VLSI prototype implementation}
 \label{sect:si:prototype}

 A $7-7-7$ network, consisting of 21 neurons and 98 synapses was fabricated in 180\,nm CMOS technology (AMS 1P6M).
 The input currents were provided through custom bias generators, optimized for sub-threshold operation \cite{delbruck201032}.
 Custom current-to-frequency converters were used to read out the outputs of neurons and send them off chip in terms of inter-event intervals.
 The weight parameters were stored on the device in latches, directly connected to the configuration lines of the synapse circuits. Custom digital logic was implemented on the chip for programming biases, weights, and monitors.
 Furthermore, the chip was connected to a PC, through a Xilinx Spartan 6 FPGA containing custom interfacing logic and a Cypress FX2 device providing a USB interface.
 Custom software routines were implemented to communicate with the chip and carry out the experiments.
 The fabricated VLSI chip was characterized through measurements as described in Sect.~\ref{sect:closedloop}, by probing individual neurons one by one.
 The measurements were repeated several times through different source and monitor neurons for each neuron to be characterized to average out mismatch effects arising from the synapse or readout circuits.
 The mean values of the measured slopes were used in a software model to train a network on the Iris flower dataset.
 The Iris dataset was randomly split into 120 and 30 samples used for training and testing, respectively.
 The resulting weight parameters were programmed into the chip and individual samples of the dataset were presented to the network in terms of currents scaled to values between 0 and 325\,nA.
 The index of the maximally active output unit was used as the output label of the network and to compute the classification accuracy.

\end{document}